\pdfoutput=1

\documentclass[11pt]{article}

\usepackage[final]{acl}

\usepackage{times}
\usepackage{latexsym}

\usepackage[T1]{fontenc}

\usepackage[utf8]{inputenc}

\usepackage{microtype}

\usepackage{inconsolata}

\usepackage{graphicx}

\usepackage{tikz}
\usetikzlibrary{shapes.geometric, arrows}
\tikzstyle{inout} = [rectangle, rounded corners, minimum width=4cm, minimum height=1cm,text centered, draw=black]
\tikzstyle{model} = [rectangle, rounded corners, minimum width=4cm, minimum height=1cm,text centered, draw=black]
\tikzstyle{layer} = [rectangle, rounded corners, minimum width=4cm, minimum height=1cm,text centered, draw=black]
\tikzstyle{arrow} = [thick,->,>=stealth]

\title{uOttawa at LegalLens-2024: Transformer-based Classification Experiments}

\author{
 \textbf{Nima Meghdadi\textsuperscript{}} and
 \textbf{Diana Inkpen\textsuperscript{}}
\\
\\
 \textsuperscript{}School of Electrical Engineering and Computer Science
\\ University of Ottawa, \\ Ottawa, ON, K1N 6N5
\\
 \small{
   \textbf{} \href{{mailto:email@domain}}{nmegh082@uottawa.ca}
   \textbf{} \href{{mailto:email@domain}}{diana.inkpen@uottawa.ca}
 }
}

\begin{document}
\maketitle
\begin{abstract}
This paper presents the methods used for LegalLens-2024 shared task, which focused on detecting legal violations within unstructured textual data and associating these violations with potentially affected individuals. The shared task included two subtasks: A) Legal Named Entity Recognition (L-NER) and B) Legal Natural Language Inference (L-NLI). For subtask A, we utilized the spaCy library, while for subtask B, we employed a combined model incorporating RoBERTa and CNN. Our results were 86.3\% in the L-NER subtask and 88.25\% in the L-NLI subtask. Overall, our paper demonstrates the effectiveness of transformer models in addressing complex tasks in the legal domain. The source code for our implementation is publicly available at https://github.com/NimaMeghdadi/uOttawa-at-LegalLens-2024-Transformer-based-Classification
\end{abstract}
\section{Introduction}

The huge amount of information and massive use of the internet has propelled to ignore legal violations, individual rights, cultural values and societal norms. These hidden violations demand serious attention and urgent solution due to serious effects on rights and justice and it requires advanced tools for professionals to effectively manage large amount of paperwork. 

Legal violation identification seeks to automatically detect legal violations within unstructured text and link these violations to potential victims. The LegalLens 2024 shared task \cite{bernsohn-etal-2024-legallens} aims to foster a legal research community by tackling two key challenges in the legal domain. Subtask A focuses on identifying legal violations (a.k.a Identification Setup) using Named Entity Recognition (NER). Subtask B focuses on linking these violations to potentially affected individuals (a.k.a Identification Setup) using Natural Language Inference (NLI).

Our team participated in both subtasks of the shared task. In subtask A, we used the spaCy library and a DeBERTa-based model. In subtask B, we developed a RoBERTa-based model combined with a CNN-based model.


\section{Related Work}
There has been extensive research on Legal Named Entity Recognition (NER) for German legal documents. \citeauthor{leitner2019fine} (\citeyear{leitner2019fine}) developed NER models using CRF and BiLSTM, while \citeauthor{darji2023german} (\citeyear{darji2023german})  used a BERT-based model.
Many languages are using NER to expedite the process of judicial decision-making. For the Turkish language, \citeauthor{ccetindaug2023named} (\citeyear{ccetindaug2023named}) developed an NER model using BiLSTM and several word embeddings like GloVe, Morph2Vec, and neural network-based character feature extraction techniques. In Portuguese, \citeauthor{bonifacio2020study} (\citeyear{bonifacio2020study}) and \citeauthor{albuquerque2023assessment} (\citeyear{albuquerque2023assessment}) focused on NER models specific to the legal domain. The former developed a model using ELMo and BERT with the LeNER-Br dataset \cite{luz2018lener}, while the latter evaluated BiLSTM+CRF and fine-tuned BERT models on legal and legislative domains to automate and accelerate tasks such as analysis, categorization, search, and summarization. 
In Italian, \citeauthor{pozzi2023named} (\citeyear{pozzi2023named}) created a model that combines transformer-based Named Entity Recognition (NER), transformer-based Named Entity Linking (NEL), and NIL prediction.
In Chinese, \citeauthor{zhang2023roberta} 
(\citeyear{zhang2023roberta}) proposed a NER method for the legal domain named RoBERTa-GlobalPointer, combining character-level and word-level feature representations using RoBERTa and Skip-Gram, which were then concatenated and scored with the GlobalPointer method. \citeauthor{lee2023learner} 
(\citeyear{lee2023learner}) also developed a legal domain NER model called LeArNER, which employs Bouma’s unsupervised learning for feature extraction and utilizes the LERT and LSTM models for sequence annotation.

\citeauthor{kim2024legal} 
(\citeyear{kim2024legal}) described methods for the COLIEE 2023 competition, using a sentence transformer model for case law retrieval and a fine-tuned DeBERTa model for legal entailment that used SNLI \cite{bowman-etal-2015-large} and MultiNLI \cite{N18-1101} datasets for training.
\citeauthor{tang2023natural} (\citeyear{tang2023natural}) explored improving legal Natural Language Inference (NLI) by employing general NLI datasets with supervised fine-tuning and examining the impact of transfer learning from Adversarial NLI to ContractNLI.
The objective of \citeauthor{valentino2024nature} (\citeyear{valentino2024nature}) is to offer a theoretically grounded characterization of explanation-based Natural Language Inference (NLI) by integrating contemporary philosophical accounts of scientific explanation with an analysis of natural language explanation corpora.
\citeauthor{gubelmann2023truth} (\citeyear{gubelmann2023truth}) investigated how large language models (LLMs) handle different pragmatic sentence types, like questions and commands, in natural language inference (NLI), highlighting the insensitivity of MNLI and its fine-tuned models to these sentence types. It developed and publicly released fine-tuning datasets to address this issue and explored ChatGPT's approach to entailment.

\section{Subtask A: Legal Named Entity Recognition(L-NER)}
Subtask A, which involves finding named entities for specific types that may appear in legal texts, is explained in this section.

We developed a BERT-based model for this subtask as part of the LegalLens task, achieving an F1-score(Macro F1-score) of 86.3\%.

\subsection{Dataset Details}
We used the dataset provided by the organizers of the shared task. The provided data was split into training and test sets, with each set consisting of tokenized text and the corresponding entities for those tokens. It is important to note that the provided test set includes labeled data, which is different from the separate test data that the organizers will use to evaluate the model. The split dataset used for validation in this research consists of 20\% of training data and is shown in Table \ref{split-ner}.

\begin{table}
  \centering
  \begin{tabular}{lc}
    \hline
    \textbf{Type} & \textbf{Number of documents} \\
    \hline
    \verb|Training|     & {568}           \\\hline
    \verb|Validation|     & {142}           \\\hline
    \verb|Test|     & {617}           \\\hline
  \end{tabular}
  \caption{The number of documents used to train the model is detailed}
  \label{split-ner}
\end{table}

The entity types are fully described in \cite{bernsohn-etal-2024-legallens}. The labels include four entity types: violation, violation by, violation on, and law, with detailed counts for each entity available in \cite{bernsohn-etal-2024-legallens}.

\subsection{Preprocessing}

For this subtask, we configure the spaCy pipeline with an emphasis on tokenization and vector initialization. The tokenizer used is the standard spaCy tokenizer, which splits the text into tokens for downstream tasks. We utilize the spacy.Tokenizer.v1 configuration, which efficiently handles tokenization according to spaCy's standards.

Next, we handle vector initialization. In this setup, vectors map tokens to high-dimensional representations, which helps capture semantic meaning during training. The data by converting the text and its annotations into a format compatible with spaCy.

\subsection{Model Training}
Our training utilizes the SpaCy pipeline configured with a transformer model and a transition-based parser for NER tasks. The deberta-v3-base model has been selected for the main transformer architecture, offering robust contextual embeddings for token-level classification \cite{he2021debertav3}.

Hyperparameters for the training are optimized based on performance on the development set. The key hyperparameters can be seen in Table~\ref{hyperparameters-ner}.

\begin{table}
  \centering
  \begin{tabular}{lc}
    \hline
    \textbf{Hyperparameter} & \textbf{Value} \\
    \hline
    \verb|Learning Rate|     & {5e-5}           \\\hline
    \verb|Batch Size|     & {16}           \\\hline
    \verb|Maximum Steps|     & {20,000}           \\\hline
    \verb|Dropout Rate|     & {0.1}           \\\hline
    \verb|Optimizer|     & {Adam}           \\\hline
  \end{tabular}
  \caption{Hyperparameters of the fine-tuned model for subtask A (L-NER)
}
  \label{hyperparameters-ner}
\end{table}

\subsection{Results and Discussion}
We found that models utilizing spaCy achieved better results compared to those without it. Additionally, BERT base models outperformed other models in our experiments. However, we discovered that initializing embeddings from Hugging Face leaderboard embeddings did not lead to improved results. Table \ref{result-ner} compares the F1-scores of various models on the labelled test data.

\begin{table}
  \centering
  \begin{tabular}{ll}
    \hline
    \textbf{Model} & \textbf{F1-score} \\
    \hline
    {roberta-base} & {52.55}\\\hline
    {nlpaueb/legal-bert-base-uncased} & {53.29}\\\hline
    {lexlms/legal-roberta-base} & {54.80}\\\hline
    {lexlms/legal-roberta-base} \\ {(Alibaba-NLP/gte-large-en-v1.5)} & {62.69}\\\hline
    {roberta-base with spacy} & {80.49}\\\hline
    {\textbf{deberta-v3-base with spacy}} & {\textbf{86.37}}\\
     \hline 
  \end{tabular}
  \caption{\label{result-ner}
    Comparison of F1-score in various models for the NER subtask
  }
\end{table}

\subsection{Direct Comparison to Related Work}
The organizers of the shared task provided a hidden test set, on which our model achieved an F1-score of 0.402, securing third place in the competition. The performance of the top five teams is presented in Table \ref{result-hidden-ner}.
\begin{table}
  \centering
  \begin{tabular}{ll}
    \hline
    \textbf{Model} & \textbf{F1-score} \\
    \hline
    {Nowj} & {\textbf{0.416}}\\\hline
    {Flawless Lawgic} & {0.402}\\\hline
    {\textit{\textbf{UOttawa}}} & {\textit{\textbf{0.402}}}\\\hline
    
    {Baseline} & {0.381}\\\hline
    {Masala-chai} & {0.380}\\\hline
    {UMLaw\&TechLab} & {0.321}\\\hline
    {Bonafide} & {0.305}\\\hline
     \hline 
  \end{tabular}
  \caption{\label{result-hidden-ner}
    Comparison of top 5 teams results on the hidden test set for the NER subtask, measured by F1-score \cite{hagag2024legallenssharedtask2024}.
  }
\end{table}

\section{Subtask B: Legal Natural Language Inference (L-NLI)}
The goal of this subtask is to automatically classify the relationships between different legal texts. Specifically, we aim to determine whether a legal premise, such as a summary of a legal complaint, entails, contradicts, or remains neutral with respect to a given hypothesis, like an online review. This task, termed Legal Natural Language Inference (L-NLI), involves sentence-pair classification to assess these relationships. By creating an NLI corpus tailored for legal documents, we facilitate applications like legal case matching and automated legal reasoning. Detailed task definitions and datasets are provided in \cite{bernsohn-etal-2024-legallens} and related resources.

\subsection{Dataset Details}
The LegalLensNLI dataset, provided by the organizers of the shared task, is specifically designed to explore the connections between legal cases and the individuals affected by them, with a particular focus on class action complaints. This dataset contains 312 entries. A comprehensive description of the dataset collection process can be found in \cite{bernsohn-etal-2024-legallens}. For this subtask, only the training set is included, and the validation set is separated into four specific domains, as outlined in Table \ref{result-domain-specific-nli}.

\subsection{Preprocessing}
This subtask has a different objective compared to Subtask A, so SpaCy may not perform well for this task. In this subtask, we began by loading the ynie/roberta-large-snli\_mnli\_fever\_anli\_R1\_R2\_R3-nli model \cite{nie-etal-2020-adversarial} using the AutoTokenizer and AutoModel classes from the transformers library. The AutoTokenizer class was employed to tokenize the input sentences, converting them into a format suitable for the roberta-large model. The tokenization process involved splitting the text into tokens and converting them into numerical representations, which are then padded and truncated to a consistent length. This ensures that the input sequences are properly aligned when fed into the model.

Following tokenization, we implemented a method to encode the combined premise and hypothesis sentences for both the Roberta model and a CNN model. The CNN model required a different form of input preparation, where the combined texts were tokenized and encoded to maintain the sequence's structure for CNN processing. These tokenized datasets were then converted into PyTorch tensors and mapped accordingly, enabling their use in a combined model that integrates both the Roberta model and the CNN. Subtask B involves finding the similarity between the hypothesis and premises. By using a CNN model to highlight the keywords in sentences, the combined model may perform better.

\subsection{Model Training}
\begin{figure}
\centering

\begin{tikzpicture}[node distance=2cm]
\node (in1) [inout] {Input Text};
\node (model1) [model,below of=in1] {Roberta Model};
\node (model2) [model,right of=model1,xshift=2cm] {CNN Model};
\node (model3) [model,below of=model1] {Concatinate Models};
\node (model4) [layer,below of=model3] {Fully Connected Layer};
\node (out1) [inout,below of=model4] {Output};

\draw [arrow] (in1) -- (model1);
\draw [arrow] (in1) -- (model2);
\draw [arrow] (model1) -- (model3);
\draw [arrow] (model2) -- (model3);
\draw [arrow] (model3) -- (model4);
\draw [arrow] (model4) -- (out1);
\end{tikzpicture}
\caption{Diagram of combined model (Roberta and CNN)} \label{fig:arch-model}
\end{figure}
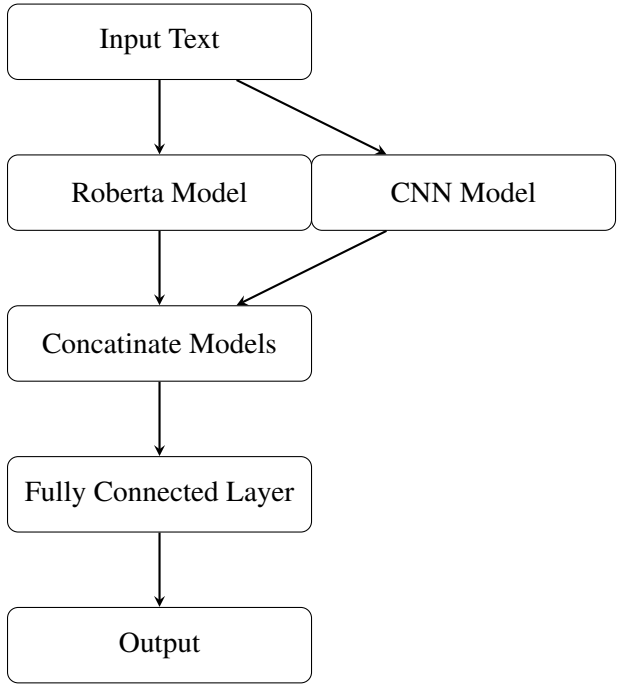

Our combined model architecture integrates the ynie/roberta-large-snli\_mnli\_fever\_anli\_R1\_R2\_R3-nli model with a custom-built CNN model for keyword detection. The Roberta model is responsible for capturing contextual information from the text, while the CNN model detects important keyword patterns within the input text. The outputs from both models are concatenated and passed through a fully connected layer (softmax) to produce the final classification decision, the architecture of the model can be seen in Figure\ref{fig:arch-model}. In more detail, the RoBERTa model consists of one embedding layer and 24 Transformer encoder layers, while the CNN model includes one embedding layer and three convolutional layers, each with a different filter size (2, 3, 4), followed by a fully connected layer. In total, we have 31 layers
Training was conducted using the Trainer class from the transformers library, which facilitated the fine-tuning of the model. We defined specific hyperparameters, that can be seen Table~\ref{hyperparameters-nli}. The model was evaluated at the end of each epoch, with the best model being saved based on the F1-score. The training process also included strategies for early stopping and warmup steps to optimize performance.

\begin{table}
  \centering
  \begin{tabular}{lc}
    \hline
    \textbf{Hyperparameter} & \textbf{Value} \\
    \hline
    \verb|Learning Rate|     & {2e-5}           \\\hline
    \verb|Batch Size (train and Eval)|     & {4}           \\\hline
    \verb|Number of Epochs|     & {20}           \\\hline
    \verb|Weight Decay:|     & {0.01}           \\\hline
  \end{tabular}
  \caption{Hyperparameters of the fine-tuned model for subtask A (L-NLI)
}
  \label{hyperparameters-nli}
\end{table}

This approach combines the strengths of both the Roberta model and CNN, allowing for a more comprehensive analysis of the text data. The fine-tuning process ensures that the model is well-suited for the specific task of classifying legal text as 'Entailed', 'Neutral', or 'Contradict'.

\subsection{Results and Discussion}
We found that pre-trained NLI models can perform significantly better than vanilla models and LLMs. Falcon 7B and RoBERTa base are the best-performing models for LLMs and vanilla models, respectively, as shown in Table \ref{result-domain-specific-nli}. The validation set has been selected to be domain-specific, based on legal\_act.

\begin{table}
  \centering
  \small
  \begin{tabular}{llllll}
    \hline
    {Model} & {CP} & { Privacy} & {TCPA} & {Wage}& \textbf{Avg}\\
    \hline
     {Falcon 7B} & {\textbf{87.2}}& {84.5}&{83.9}&{68.5}&{81.02}\\
     {without cnn} & {87.23} & {85.48}&{83.88}&{90.6}&{86.77}    \\
     {roberta-base} & {82.9} & {62.0}&{69.5}&{69.7}&{71.02}    \\
     {Our model} & {84.4} & {\textbf{90}}&{\textbf{84}}&{\textbf{96}}  &{\textbf{88.6}}   \\
     \hline
  \end{tabular}
  \caption{\label{result-domain-specific-nli}
    Comparison of F1-score on the validation set for various models for the NLI task for specific-domain (Consumer Protection, Privacy,TCPA and Wage)
  }
\end{table}

\subsection{Direct Comparison to Related Work}
The shared task organizers evaluated the models using a hidden test set, where our model attained an F1-score of 0.724, placing fifth in the competition. The results of the top five teams are detailed in Table \ref{result-hidden-nli}.

\begin{table}
  \centering
  \begin{tabular}{ll}
    \hline
    \textbf{Model} & \textbf{F1-score} \\
    \hline
    {1-800-Shared-Tasks} & {\textbf{0.853}}\\\hline
    {Baseline} & {0.807}\\\hline
    {Semantists} & {0.785}\\\hline
    {Nowj} & {0.746}\\\hline
    {\textit{\textbf{UOttawa}}} & {\textit{\textbf{0.724}}}\\\hline
     \hline 
  \end{tabular}
  \caption{\label{result-hidden-nli}
    Performance of the leading 5 teams on the hidden test set in the NLI subtask, measured by F1-score \cite{hagag2024legallenssharedtask2024}.
  }
\end{table}

\section{Conclusion and Future Work}
Our experiments demonstrated the success of transformer models, such as RoBERTa and DeBERTa, in handling complex legal tasks, including violation detection and inference. In Subtask A (L-NER), incorporating DeBERTa into the spaCy pipeline yielded strong results for legal named entity recognition. In Subtask B (L-NLI), combining RoBERTa with CNN for keyword detection boosted classification accuracy.

However, despite using robust models, generalizing to unseen cases proved challenging, particularly with nuanced legal language. While the CNN improved phrase detection, more advanced methods, like attention mechanisms, may further enhance performance.

Future work should explore architectures fine-tuned on legal texts or combine transformers with graph models to capture legal relationships. Additionally, leveraging LLMs like GPT-4 could improve legal reasoning.

\bibliography{costum}



\end{document}